\documentclass[10pt,conference,a4paper]{IEEEtran}
\usepackage{times,amsmath,epsfig}
\title{Cortex simulation system proposal using distributed computer network environments}
\author{%
{Boris Toma\v{s}{\small $~^{\#1}$} }%
\vspace{1.6mm}\\
\fontsize{10}{10}\selectfont\itshape
$^{\#}$\,University of Zagreb, Faculty of Organization and Informatics, Pavlinska 2, Vara\v{z}din, Croatia\\
\fontsize{9}{9}\selectfont\ttfamily\upshape
$^{1}$\,boris.tomas@foi.hr\\
\vspace{1.2mm}\\
\fontsize{10}{10}\selectfont\rmfamily\itshape
\fontsize{9}{9}\selectfont\ttfamily\upshape
}

\begin{document}
\maketitle
%


\begin{abstract}
In the dawn of computer science and the eve of neuroscience we participate in rebirth of neuroscience due to new technology that allows us to deeply and precisely explore whole new world that dwells in our brains. This review paper is merely insight to what is currently ongoing research in the interdisciplinary field of neuroscience, computer science, and cognitive science.
\end{abstract}

\maketitle

\section{Introduction}

There are many new and different approaches to the understanding of human brain, some use neuroscience/neurobiology to investigate biological characteristic of working brain. Those characteristics are well researched up to the level of molecules that form neurons and inter neuron connections – synapses like described in \cite{Rossmann1996}
\\
Other approach for understanding human brain is creating simulations or even emulations of human brain using current achievements in the field of computer science. Those simulations use tremendous amount resources to simulate elements of human brain; simulations even go up to the level of simulating movement of single molecule. By this day there has not been successful simulation of entire human brain.\cite{Hemmen2006} However, there are simulations of single columns of a human brains and full neural system simulation of lesser life forms like worm \textit{Caenorhabditis elegans} that is a part of Open worm project\cite{Williams2011}. Currently there has been some advances in simulating rat's brain.\cite{Mao2011}
\\
Important approach for this review is artificial intelligence (AI) which uses several techniques to implement some sort of artificial intelligence. These systems are used for decision making in financial institutions, they are being used in computer games for simulating opponents, and many more applications. AI uses many different approaches, one of it is neural networks of artificial neurons\cite{Braspenning1995}. Others are complex expert systems that are capable of learning. However, AI research outputs agents that are specialized for resolving only specific problem for example AI using neural network for ballistic interception movement\cite{Roisenberg1996}.
\\
By this day there is no know AI agent that is being self-aware and intelligent enough to confront human intelligence\cite{Andras2005}. However, there are intelligent systems, for example Watson that is miraculous piece of software that is beyond of human species regarding data processing, data mining, language recognition, processing and finally conclusion making\cite{Ferrucci2012}. This system only possess huge amount of data, and it relays on the data. In elementary school we are taught that intelligence is not equal to knowledge but this is ability of an entity how well it manages in new and unknown situations.\cite{Sternberg2000} If you cut power electricity of Watson machine "he" will do nothing about it, next time “he” will simply reboot and will not complain or regret about being “plugged out” thus cease to exist.
\section{State of the art}

\subsection{Computer science}
The most relevant field of science for this research is computer science, especially areas of software development, software architecture, networking, AI, computing, grid computing, distributed computing, parallel computing. It is known that neural networks are used in decision making, for example in stock markets\cite{Dase2010}. Current advancements in this field are very oriented towards better computing capabilities of neural networks, to provide better and faster results for decision makers. As being invented in late 1960\cite{Solomonoff1966} when computer power and our knowledge of working brain was not know enough, it can be easy assumed that this approach should be altered by incorporating more important concepts of human brain and it should be implemented using more advanced computer technologies as previously mentioned e.g. parallel computing, distributed or grid computing.
\subsection{Neuroscience}
Advancements in brain architecture, neuron architecture, neuroplasticity, connectome concepts\cite{Sporns2005}, motoric related formation of neural networks, emphatic related formation of consciousness and many more are relevant areas of neuroscience. Although, many advancements in neuroscience have been made, there still are many unsolved problems relevant to this work\cite{Hemmen2006}:
\begin{itemize}
\item How is consciousness formed?
\item Where is memory stored and how it is retrieved?
\item How does the brain transfer sensory information into coherent, private percepts?
\item How are the senses integrated?
\item How does previous experience alter perception and behaviour?
\end{itemize}

\subsection{Cognitive and other sciences}
Swarm theory\cite{Zhu2010} uses examples from nature (ants, termites, birds,…) to enhance computing techniques in computer science, again, to solve problems for e.g. decision makers., cognitive models, consciousness tests, consciousness models. Latest neural networks achievements thrive to understand how neuron can be simulated in order to provide better understanding of neuron and more importantly to improve computing and decision making. There is also hardware approach\cite{Bo2000} that tries to implement concept of biological neuron on a microscopic level of integrated circuits. By this day, there are electronic chips that can communicate with real and living neurons in brain, this is one direction - human computer interfacing\cite{Bahari2010}; other is to create computer chips that process information the same way this information is processed in real biological neural networks
\section{Relevant projects}
There are several project that are relevant to this work:
\subsection{Blue Brain Project\cite{Markram2006}}
The Blue Brain project represents an essential first step toward achieving a complete virtual human brain. The researchers have demonstrated the validity of their method by developing a realistic model of a rat cortical column, consisting of about 10,000 neurons.
Blue Brain Project is collaboration between several universities in the world and industrial partners like IBM. Project goal is the simulation of human brain, not achieving an artificial intelligence in machines, however scientist do expect that some form of consciousness or intelligence might emerge. This project uses huge amount of resources from IBM supercomputers. Resources for each neuron are equal to single everyday laptop. This demand for resources come from fact that this project simulates every chemical characteristic of neuron. 
\\
In this paper () such use of resources is not necessary because not all neuron features are required for achieving intelligence, some features are required for biological achievability and sustainability.
\subsection{Cajal Blue Brain Project\cite{UniversidadPolitecnicadeMadrid2010}}
This project is run by several Spanish institutions, it is a part of Blue Brain Project but has taken a different path. As Blue Brain Project it also thrives to simulate portions of human brain using software development on powerful supercomputers.
\subsection{Nengo\cite{Sarangapani2005}\cite{CentreforTheoreticalNeuroscience}}
This project is carried out by Canadian scientists. Nengo is a software for simulating large-scale neural network systems. Largest simulation is carried out on super computers and is called Spaun. In Nengo, network architecture is designed in really nice and easy GUI.
Positive thing is that Nengo project provides the visual input for the simulated brain: 28 x 28 pixels. Also they provide output for the simulated brain in the form of robotic arm. Real problem is that this hand is nothing less than a read only output, it could easily be a computer screen. There still does not exit correlation between input and output, like any dependence, something that enables Spaun to change the outside environment and that way influence its own state and conditions.
Spaun neuron model, and network model has to be, in beginning set to solve specific problem. This means that when simulation starts it reads predefined values. This shows large intrusion in self determination of the network. In this work it is intended to simplify the model as much as possible, and let the neuron model organize its connections and in indirectly the network architecture. 
\subsection{Would digital brain have a mind?\cite{Holmes2002}}
Nevile Holmes analyses reason why digital brain still does not have mind of its own or consciousness. Those reasons are:
\begin{itemize}
\item The activation of a classical neuron is not an arithmetic sum of the synaptic effects. Rather, a complex process involving the intervals between action potentials at individual synapses and their relative timings between synapses deter- mines the activation. The more neurons are studied the more such complexity is revealed.
\item The human nervous system contains many different kinds of neurons and many kinds of glial cells. The glial cells provide more than support because they signal and have synapses just as neurons do.
\item Action potentials alone do not control nervous signalling. Graded potentials and hormonal signalling also play a part, as does the great variety of different neurotransmitters and hormones.
It is noticeable that author recognizes lack of biological importance to existing artificial neuron model.
\end{itemize}
\section{Research questions}
Research has found that there is no artificial intelligence agent that is comparable to human intelligence\cite{Hendler2006}. Also, it is possible that some concepts from swarm theory can be used to explain formation of intelligence and consciousness in our brains. It is assumed, by some authors, that consciousness is a by-product of evolution process\cite{Berwick2013}. This knowledge can be used to create similar concepts in software.
Swarm theory simply determines properties and functions, it does not explain why swarm intelligence emerges. This is hard to prove but not impossible. 
\\There is common feature found in every swarm theory example – it is “the purpose” – there has to be a reason why entities form a swarm, and because of this reason swarm thrives to fulfil this reason more efficiently than the single swarm entity. The same principle can be applied to human brain and biological neural networks where our neurons form swarms with its functions (e.g. transferring action potential between neurons), but they need a reason to achieve swarm intelligence: that may be overall human intelligence with purpose to live and to survive. This is relatively philosophical issue that has been very well elaborated. Existing AI, neural networks, brain simulations do have a purpose: computing data, data mining, simulating intelligence in computer games, simulating brain activity etc. but none has the purpose to live, to survive or to become better. As by this date not AI agents or software in general has passed the Turing test, one of the famous tests is being held by Loebner test each year. Most "human like" AI agent receives an award, no agent has received maximum grade in this competition.\cite{Saygin2001}
It believed that it is possible to implement such simple software nodes that have ability to communicate together (inside single computer node and over the networks like Internet). This single node will resemble to single artificial neuron in artificial neural networks, but it would not be the same, it will resemble more to biological neuron.  That is, it will implement some newly discovered concepts from biological brains (neuroplasticity, temporal computing…). On the other hand, purpose would not be to make simulation of living brain because simulating whole brain and its concepts require huge amount of resource. To minimize eventual use of large amount of resources intention is to include only significant features of living brain, for example not all brain features are required for forming of consciousness and intelligence, precisely, they are required in biological and chemical sense: for example synapses are one important biological feature but one of its main purposes is to make directional path between two neural cells – to isolate action potential propagation. This is not required in software because directions can be created differently like using linked lists or other pointer usages.
Network architecture is not relevant for artificial networks, but it is for biological neural networks. Only important feature in neural networks is that they can encode information as unique sequence of neurons, something very similar to hash functions. 
To prevent overload of data, thus lack of space to store data some processing should be done to purge irrelevant data, in biological sense this is called forgetting, we forget simply to purge our brain of not necessary information, along with forgetting some scientist find dreams ways of purging irrelevant data and confirming data that is important.\cite{Lindsley1989}
Interesting questions is why does interaction between encoded information (in node sequences) emerges intelligence and/or consciousness?
There is apparently huge gap between neuroscience and artificial intelligence because in 1960’s that gap did not existed, researchers created artificial neuron based on current knowledge of biological neuron. As the years have passed by, more neuroscience discoveries were made and we now know much more than 50 years ago, on the other hand artificial neurons and artificial neural networks (ANN) went in directions of computing and data processing. Over time this gap became significant. However, there are attempts to create simulations/emulations, later in further sections. 
General purpose of this work is to try to fill this gap by designing new artificial neuron model with selective characteristic (of biological neuron) that can coexist in modern computer systems and is capable of communicating in computer networks like Internet.
\begin{enumerate}
\item Is it possible to create model of artificial neuron (node) that is capable of network communication (using custom protocol)?
\item Is it possible to create such system and network architecture that can sustainably host artificial network of nodes?
\item Is it possible to create such system that shows self-consciousness and intelligence that can satisfy relevant tests?
\item Is it possible to create a form of swarm intelligence artificially using network of simple nodes?
\end{enumerate}

\section{Approach}
Using extensive literature research from all fields, it is intend to find out needed characteristics and features that are necessary to be incorporated in a new neural model. After designing the model it is planned to start experiment phase:
\subsection{Development stage} 
During this stage new model will become implemented as a single piece of software. Also, it is planned to take such approach that will not be as much model oriented but some form of wrapping technology around the model. This is necessary because eventual easier model change in real case scenarios during real-time.
\subsection{Development of new networking protocol}
Design of necessary network protocol that can support communications between nodes. This protocol should be as light as possible, and even as low level implementation as possible on the ISO/OSI network stack.
\subsection{Development of system's input and output}
Input/output is technology that allows system to interact with its surroundings and environment.
\\
During experimental phase it is intend to analyse and explain all behaviour that may arrive correlating it to related existing discoveries in neuroscience and cognitive science.
After experimental stage there is testing stage, it is planned to use existing tests to prove or disapprove existence of any form of intelligence and consciousness.
\section{Research design \& methodology}
Plan is to partially follow \textit{constructive research method}:
\begin{itemize}

\item{Fuzzy information sources}\\
In proposed fields there are sources like science paper, conference articles, books, book chapters. However, there are also popular articles that cannot be taken as reliable. As the nature of this work is highly experimental, those sources may provide fuzzy guidelines for research and next steps.
\item{Theoretical body of knowledge}\\
Using gathered sources it is planned to determine a strong body of knowledge.
\item{Relevant problem definition}\\
Relevant problem is initially described in previous sections, during definition of theoretical body there might be changes in definition of relevant problem definition.
\item{Solution design}\\
One of results of this research is a system that uses software and network infrastructure to prove or disapprove hypothesises.
\item{Practical and theoretical relevance}\\
Finally there is extensive analysis of experimental solution. In concluding sections there should be recognized relevant influence on theoretical fields as well as new practical uses of designed experimental system.
\end{itemize}

\section{Significance}
Success of this research would be that it can prove that life as we know it can exist even in software environments (although, in 2011 scientists from NASA made a breakthrough and found life form that is not phosphorous based as the life we know it). If successful it can explain how and why consciousness forms in living brain, it can certainly help dealing with many neural diseases.
If unsuccessful research can pinpoint potential pitfalls in research for true artificial intelligence.

\section{Possible pitfalls}
\begin{itemize}
\item{Complexity}\\
System complexity might become pitfall if it implements too much characteristics from biological examples.
\item{Resources}\\
System is very resources dependent and use of European research computing grids would be required. To achieve distributed network topology it would be necessary to deploy system nodes on remote locations over the Internet, population can help in this research by assigning certain amount of resources to this research (similar approach as SETI@Home)
\item{Related work and literature}\\
There are significant resources in each field, as this research is in several fields there is substantial literature. However, in interdisciplinary field there is not much resources but few really significant projects that deal with the similar issue.
\end{itemize}
\section{Conclusion}
This work is highly experimental and includes cutting edge of different science fields. There is also a great risk that it would not be possible to prove hypothesis. Exploring our mind is something humankind is trying to do since ancient times. Today technology to scan and monitor non invasive, is present and can be used to monitor activity inside brain. Neuroscience has gained giant leap towards understanding biology of our brains. However, brain is data/signal processing device that behaves more as computer, this interdisciplinary nature of this work is something currently being on the cutting edge of science. Countries, philanthropists, and many more invest great amounts of resources into this field, because it is recognized that brain is one universe we yet need to explore.

\hfill December 12, 2013

\ifCLASSOPTIONcaptionsoff
  \newpage
\fi

\bibliography{btcol}
\end{document}